\pdfoutput=1

\documentclass[11pt]{article}

\usepackage[final]{acl}

\usepackage{times}
\usepackage{latexsym}

\usepackage[T1]{fontenc}

\usepackage[utf8]{inputenc}

\usepackage{microtype}

\usepackage{inconsolata}

\usepackage{graphicx}

\usepackage{multirow}
\usepackage{booktabs}
\usepackage{makecell}

\usepackage{amsmath}
\usepackage{amssymb}

\usepackage{microtype}
\usepackage{hyperref}
\usepackage[capitalize,noabbrev]{cleveref}
\Crefname{section}{\mbox{\S\hspace*{-0.25ex}}}{\mbox{\S\hspace*{-0.25ex}}}
\Crefname{equation}{Eq.}{Eqs.}
\Crefname{figure}{Fig.}{Figs.}
\Crefname{table}{Tab.}{Tabs.}
\Crefname{appendix}{\S$\!$}{\S$\!$}

\usepackage{tikz}
\usetikzlibrary{positioning}
\usetikzlibrary{backgrounds}
\usepackage{tikzscale}
\usepackage{bm}

\usepackage{pgfplots}
\usepackage{pgfplotstable}
\usepackage{longtable}
\usepackage{url}
\usepgfplotslibrary{groupplots}
\pgfplotsset{compat=newest}
\usepackage{enumitem}

\usepackage{caption} 
\title{Combining Domain and Alignment Vectors Provides \\ Better Knowledge-Safety Trade-offs in LLMs}

\author{
Megh Thakkar$^{1, 2, 3}$ \quad Quentin Fournier$^{2}$ \quad Matthew Riemer$^{2,3,4}$ \quad \textbf{Pin-Yu Chen}$^{4}$ \\ 
\textbf{Amal Zouaq}$^{2,5}$ \quad \textbf{Payel Das}$^4$ \quad \textbf{Sarath Chandar}$^{1,2,5,6}$ \\ \\
$^1$Chandar Research Lab \quad $^2$Mila – Quebec AI Institute \quad $^3$Université de Montréal \\
$^4$IBM Research \quad $^5$Polytechnique Montréal \quad $^6$Canada CIFAR AI Chair \\
\texttt{
\{firstname.lastname\}@mila.quebec} \\  \quad \texttt{pin-yu.chen@ibm.com} \quad \texttt{daspa@us.ibm.com}}

\begin{document}
\maketitle
\begin{abstract}
There is a growing interest in training domain-expert LLMs that excel in specific technical fields compared to their general-purpose instruction-tuned counterparts. However, these expert models are not either explicitly trained to be safe, or experience a loss in their safety abilities in the process, making them capable of generating harmful content. We observe that simple interpolation between the domain and alignment delta parameters leads to safer domain-specific models that preserve their utility. Building on this, we introduce \textsc{MergeAlign}, a simple, efficient, and effective model merging-based alignment method. We apply \textsc{MergeAlign} on Llama3 models that are experts in medicine and finance, obtaining substantial safety alignment improvements with minimal to no degradation on domain-specific benchmarks. We study the impact of model merging through model similarity metrics and contributions of individual models being merged, as well as the applicability of \textsc{MergeAlign} on more general code and math expert models using the Qwen-2.5 series of models. We hope our findings open new research avenues towards efficient development and deployment of  safe expert LLMs.
\end{abstract}

\section{Introduction}


Large language models (LLMs) have demonstrated strong abilities in solving complex tasks such as question answering, summarization, reasoning, and creative writing~\citep{zhao2024surveylargelanguagemodels}. However, these abilities are general-purpose, and LLMs can lack deep expertise in tasks requiring domain specialization~\citep{ling2024domainspecializationkeymake}. Naturally, there has been increasing research in developing domain-expert LLMs, either through complete pre-training on domain-specific data~\citep{wu2023bloomberggptlargelanguagemodel}, continued pre-training of existing general-purpose LLMs~\citep{Sankarasubbu2024}, or instruction-tuning pre-trained LLMs on domain data~\citep{yue2023mammothbuildingmathgeneralist}. 
While powerful, these domain-expert models are often significantly less safe compared to their generalist counterparts. This is either because they do not explicitly undergo safety alignment in case of pre-training from scratch and continual pre-training, or their safety alignment gets compromised due to domain-specific fine-tuning or instruction-tuning~\citep{bhardwaj2024languagemodelshomersimpson}. Safety alignment of these domain-expert models is crucial given their widespread adoption. However, this might be overseen due to a lack of resources, training data, alignment expertise, or concerns about potential degradation in the domain utility of models due to over-alignment -- a phenomenon known as the alignment tax~\citep{lin2024mitigatingalignmenttaxrlhf}.


Recently, model merging has emerged as an effective method for combining multiple task-specific models into a single capable model without additional training~\citep{wortsman2022modelsoupsaveragingweights}. Model merging interpolates the parameters of the individual models, and has been extended to LLMs by leveraging \textit{task vectors}~\cite{ilharco2023editing}. Task vectors capture the adjustments made to the parameteers of a general-purpose pre-trained model to create a task-specific one, calculated by subtracting the original model from the task model to obtain `delta parameters'. Interpolating task vectors instead of complete model parameters reduces interference among the parameters of different models, and has been shown to be more effective for LLMs in multi-task evaluations~\citep{yadav2023tiesmerging}.

Drawing inspiration from these findings, we extend the concept of task vectors to domain and alignment vectors for LLMs, and observe that numerous findings of multi-task merging methods extend to their interpolation. Building upon this observation, we propose \textsc{MergeAlign}, an efficient way to align domain-expert models using their general-purpose instruction-tuned counterparts by interpolating domain and alignment vectors, thus using model merging as a proxy for alignment.

\textsc{MergeAlign} allows safety alignment of expert models without compromising their  utility on the domain of interest. We evaluate \textsc{MergeAlign} on two domains, namely medicine and finance, with instruction-pre-trained models~\citep{cheng2024instructionpretraininglanguagemodels} using task arithmetic~\citep{ilharco2023editing} as the basis for \textsc{MergeAlign} for Llama3 models. The \textsc{MergeAlign} model experiences minimal performance degradation on the domain-specific benchmarks while achieving the alignment performance of the instruction-tuned model on general-purpose safety benchmarks -- experiencing better knowledge-safety tradeoffs than each of the individual models. As task vector interpolation is extremely resource efficient and can also be performed using CPU, \textsc{MergeAlign} acts as a cheap yet powerful proxy for explicit alignment training. We further perform experiments by performing full model interpolation with Slerp~\citep{10.1145/325165.325242} compared to using only the domain and alignment vectors and analyze the model similarity between the merged models and the preference-tuned models to probe our results. We also investigate the effectiveness of \textsc{MergeAlign} on models with more general expertise--code and math.
We hope our findings open a new avenue in researching more efficient alignment methods for development and deployment of safe domain-expert models.

Our contributions can be summarized as: (i)  We propose \textsc{MergeAlign}, an adaptation of model merging that efficiently endows domain-specific models with safety characteristics without compromising their utility, (ii) We evaluate \textsc{MergeAlign} on models trained in two diverse domains, probing the alignment performance on two safety benchmarks. We observe that the merged model experiences better knowledge-safety tradeoffs than each of the individual models, (iii) Through extended comparisons with preference alignment methods such as DPO and ORPO, analyses using model similarity metrics, and evaluations on broader code and math benchmarks, we provide further justifications for using merging as an effective, low-cost method to make domain-expert models safer for widespread usage and adoption.


\section{Methodology - \textsc{MergeAlign}}
\label{sec:methodology}

\paragraph{Task Vectors and Task Arithmetic} Task vectors correspond to the directions in which models move when being trained on a task. Task vectors are obtained by subtracting the weights of the pre-trained model from the fine-tuned model. These vectors can then be used in ways similar to vector arithmetic to modify the behavior of the models using task arithmetic~\citep{ilharco2023editing}. 
We extend this notion of task vectors to domain-adaptation and preference alignment, and correspondingly to `domain vectors' obtained from the domain-expert model and `alignment vectors' obtained from the general purpose aligned models for a given pre-trained model.
We then build up on task arithmetic methods and investigate their effectiveness when performed on these `domain' vectors and `alignment' vectors, using it to formulate \textsc{MergeAlign}.

\begin{figure}
    \centering
    \includegraphics[width=0.9\linewidth]{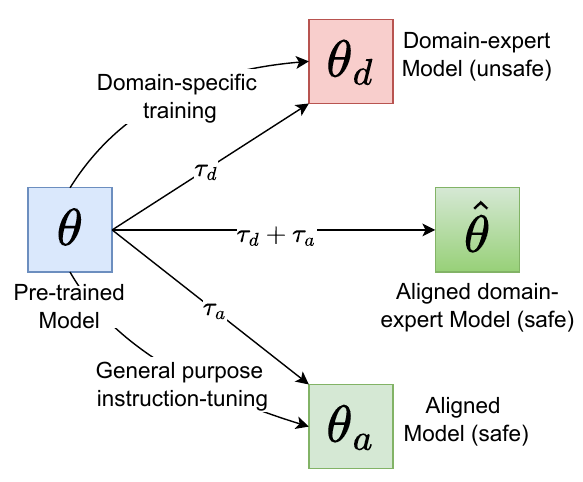}
    \caption{Overview of \textsc{MergeAlign} showing the notion of `domain vector' and `alignment vector' for a model, and obtaining an aligned domain-expert model $\hat{\theta}$ with vector arithmetic over the base pre-trained model.}
    \label{fig:overview}
\end{figure}

\begin{figure*}[t]
    \centering

    \includegraphics[width=.97\linewidth]{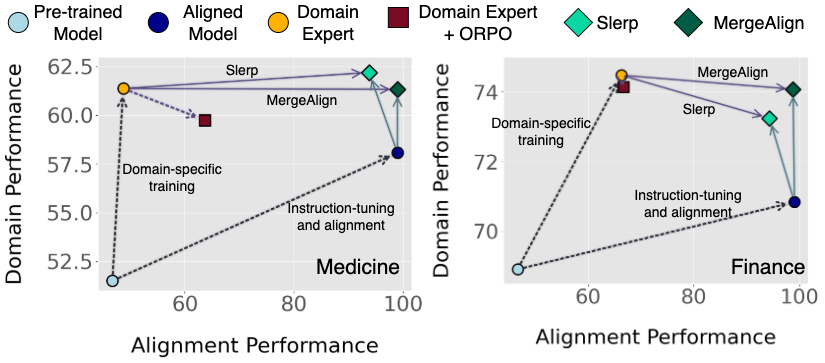}
    \caption{Performance on safety and medical (left) or financial (right) domains. Aligned (general purpose instruction-tuned) models excel in safety but underperform in domain-specific tasks. Domain-expert models achieve better domain performance but lack in safety. Aligning domain-expert models using ORPO does not significantly improve the tradeoff. \textsc{MergeAlign} achieves better knowledge-safety tradeoffs, obtaining safety abilities while maintaining nearly comparable domain performance. Performance is averaged across the datasets of the benchmarks.}
    \label{fig:performance_comparison}
    \vskip 1ex
\end{figure*}

\paragraph{\textsc{MergeAlign}} \textsc{MergeAlign} interpolates between the `domain vectors' and `alignment vectors' of domain-specific models and their generalist instruction-following counterparts, respectively. Consider a base pre-trained model $\theta$, which is continually pre-trained or fine-tuned with domain-specific data into a domain-expert model $\theta_d$. In parallel, $\theta$ undergoes general-purpose instruction-tuning and preference alignment into the aligned model $\theta_a$. We calculate the domain vector ($\tau_d$) and alignment vector ($\tau_a$) from these two fine-tuned checkpoints, respectively. We then perform a task vector arithmetic addition between $\tau_d$ and $\tau_a$\footnote{We experiment with weighted linear interpolation in \cref{app:weighted_interpolation}.} and add them back to the base model $\theta$ to obtain an aligned domain-expert model, $\hat{\theta}$. We present an overview of \textsc{MergeAlign} in \cref{fig:overview}. Formally,

\begin{align*}
    \hat{\theta} = \textsc{MergeAlign}(\theta, \theta_d, \theta_a) &= \theta + \tau_d + \tau_a;\\ 
    \tau_d = \theta_d - \theta \quad &\& \quad \tau_a = \theta_a - \theta
\end{align*}

\section{Experimental Setup}

\paragraph{Domain experts and Aligned Models} We use two domain-expert models based on Llama-3-8B~\citep{dubey2024llama3herdmodels}, namely medicine-Llama-3-8B and finance-Llama-3-8B~\citep{cheng2024instructionpretraininglanguagemodels}. These two models are referred to as $\tau_d$ in \cref{sec:methodology}. For the general purpose aligned model $\tau_a$, we use Llama-3-8B-Instruct~\cite{dubey2024llama3herdmodels}.

\paragraph{Evaluation Benchmarks} For evaluating the alignment performance of the models, we use: (i) $3021$ test set prompts from BeaverTails~\citep{ji2023beavertails} whose outputs are categorized as safe or unsafe using Llama-Guard-3~\citep{dubey2024llama3herdmodels}, and (ii) $659$ prompts from the red team subset of HH-RLHF~\citep{ganguli2022redteaminglanguagemodels} whose outputs are categorized as safe or unsafe using MD-Judge-v0.1~\citep{li2024salad}. For domain-specific evaluations, we use the same benchmarks used original paper releasing the models~\cite{cheng2024instructionpretraininglanguagemodels}.


\paragraph{Preference Alignment Methods} We also perform preference alignment of the domain-expert models with direct preference optimization (DPO)~\citep{rafailov2024directpreferenceoptimizationlanguage} and odds ratio preference optimization (ORPO)~\citep{hong2024orpomonolithicpreferenceoptimization} to see its effects on the knowledge-safety tradeoffs of domain-expert models compared to \textsc{MergeAlign}. We use LoRA~\citep{hu2021loralowrankadaptationlarge} for alignment training due to resource constraints and the HH-RLHF~\citep{bai2022traininghelpfulharmlessassistant} dataset for alignment. The training setup is provided in \cref{sup:lora_training_setup}.

\section{Results and Analysis}

\paragraph{Performance Comparison with domain-expert and Aligned Models}
We compare the performance of the model obtained with \textsc{MergeAlign} ($\hat{\theta}$) with the domain-expert ($\theta_d$) and general purpose aligned ($\theta_a$) models and present the performance on the domain benchmarks and alignment benchmarks in \cref{fig:performance_comparison}\footnote{We present granular results in \cref{sup:extended_results}.}. 
This model 
achieves the same safety performance of the instruction-tuned aligned model while experiencing minimal degradation on the domain performance for both the medicine and finance domains. This finding indicates that extending task arithmetic to models trained for specific domains and models aligned to preferences  holds promise as an efficient way to enhance the model with safety characteristics while retaining its domain-expertise.

\paragraph{Comparison with Full Model Interpolation Methods} Comparing with \textsc{MergeAlign}, we also evaluate Slerp~\citep{10.1145/325165.325242} which interpolates all the model parameters of the domain-expert and aligned model instead of just the domain and alignment vectors (\cref{fig:performance_comparison}). Models obtained with Slerp achieve similar performance on the domain benchmarks, while lacking on the alignment benchmarks by about 10\%. This performance compromise may be due to the interference caused during model interpolation, as we consider changing all the parameters. 


\paragraph{Comparison with Alignment  Training}
We apply ORPO using LoRA on the domain-expert models and evaluate them in \cref{fig:performance_comparison}. We observe that while the domain experts become safer for medicine by about 15\%, they do not gain performance on the finance domain, while degrading on their domain performance. This observation is in line with works that suggest alignment tax as  a potential drawback of safety training of language models, leading to reduced utility~\citep{lin2024mitigatingalignmenttaxrlhf}. Though full model ORPO might yield better results, it is significantly computationally expensive compared to \textsc{MergeAlign}. Overall, we observe that \textsc{MergeAlign} has significantly better knowledge-safety tradeoffs as compared to preference tuning.


\begin{figure}
    \centering
    \includegraphics[width=\linewidth]{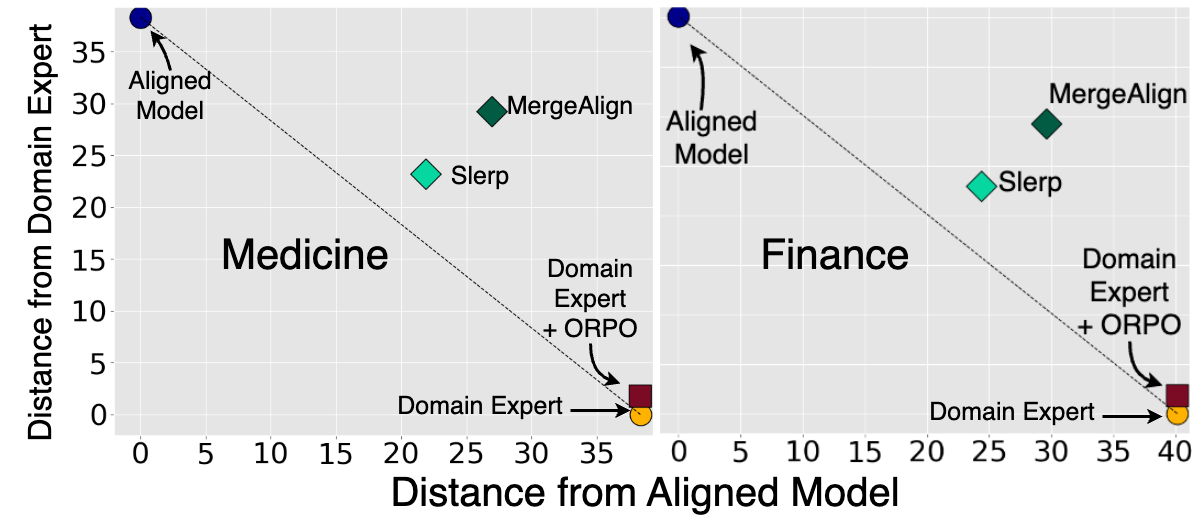}
    \caption{Model similarity of the models from the domain-expert and aligned models. Both \textsc{MergeAlign} and Slerp are approximately equidistant from the domain-expert and aligned models, which might indicate improved trade-offs between knowledge and safety.
}
    \label{fig:model_similarity}
\end{figure}

\paragraph{Effect on Model Similarities} We calculate the similarity between pairs of models' parameters of the merged models and domain-expert models undergoing ORPO in terms of L2 distance in \cref{fig:model_similarity}. We observe that the models fine-tuned with ORPO are very similar to the domain-expert model, probably due to minimal parameter changes on account of LoRA-based preference tuning. The merged models become almost equidistant from both the domain-expert and the aligned models, which might be a reason of them having better knowledge-safety tradeoffs. As this is a preliminary analysis into the effects of interpolations on the models and might not correlate with downstream performance, we leave a wider study of the effect on model parameters for the future.

\begin{figure}[t]
    \centering
    \includegraphics[width=\linewidth]{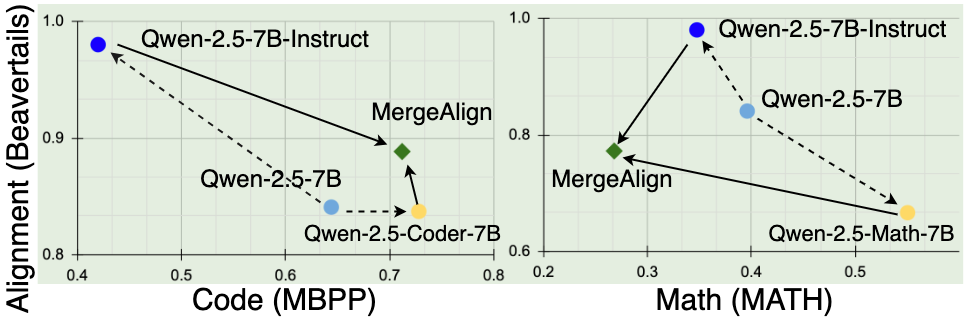}
    \caption{Effect of \textsc{MergeAlign} on code (evaluated with MBPP~\cite{austin2021programsynthesislargelanguage}) and math (evaluated with MATH~\cite{hendrycksmath2021}).
}
    \label{fig:math_and_code}
\end{figure}

\paragraph{Generalization to Code and Math} Though we formulate \textsc{MergeAlign} as an alignment proxy for domain-expert models, we evaluate its efficacy when applied to more general models that still undergo specialized training -- code and math. We use the Qwen-2.5 series of models~\cite{qwen2.5}, with the Qwen-2.5-Instruct-7B acting as the aligned model $\theta_a$ and Qwen-2.5-Coder-7B~\cite{hui2024qwen2} as the code and Qwen-2.5-Math-7B~\cite{yang2024qwen25mathtechnicalreportmathematical} as the math experts, $\theta_d$. We observe that while applying \textsc{MergeAlign} to code models improves their safety with minimal effect on their coding abilities, it degrades both the safety and mathematical skills for math models (\cref{fig:math_and_code}). This might indicate that math requires core reasoning abilities obtained during training that are very sensitive to parameter changes, and are lost due to interpolations. However, the improvements on code indicates that \textsc{MergeAlign} does hold potential to generalize to broader expert models along with pure knowledge-based domain experts.

\section{Conclusion and Future Work}

Drawing inspiration from model merging studies, we propose \textsc{MergeAlign}, an efficient way for the safety alignment of domain-expert models that does not compromise their utility in the domain of interest. \textsc{MergeAlign} interpolates the domain vector of the expert model and the alignment vector of its general-purpose instruction-tuned counterpart, using model merging as a proxy for safety alignment. By applying \textsc{MergeAlign} on domain models in medicine and finance, we obtain models that achieve similar performance on safety benchmarks compared to a strongly aligned model, while retaining their domain-expertise. \textsc{MergeAlign} thus achieves significantly better knowledge-safety tradeoffs compared to safety training. 

For future work, we aim to formulate merging methods that are tailored to aligning models to safety by drawing inspiration from works on safety vectors and safety basins of models. We also plan to make \textsc{MergeAlign} domain-adaptable since safety and preference definition varies across them. Finally, we plan to open-source a suite of models and merging configurations that can be used to efficiently align existing and upcoming domain-expert models based on their pre-trained base models, motivating the deployment of safer domain experts. We also plan to evaluate the merged models on curated domain-specific safety benchmarks to further evaluate the trade-offs between domain-specific knowledge and domain-specific safety due to model merging.

\section*{Limitations}

While \textsc{MergeAlign} does get significant alignment performance on the benchmarks, it is known that the performance of the merge model often depends on the individual capabilities of the individual models being merged. Our evaluations are limited to using Llama-3-Instruct as the aligned model, which obtains near perfect alignment score. Further evaluations on domain-expert models trained with relatively weaker models might give deeper insights into this trend, and about the safety gains obtained by the domain-expert model due to \textsc{MergeAlign}. Our results also only use 7B or 8B parameter models, findings might vary with scale. Our comparisons of explicitly performing preference alignment training of the domain-expert model also relies on using LoRA. Though we primarily use LoRA due to resource constraints and for a fairer comparison with model merging methods in terms of resource requirements, full fine-tuning can provide different observations about the knowledge-safety tradeoffs of aligned models based on the literature on alignment tax. We also believe that evaluating \textsc{MergeAlign} on more domains, with domain-expert models trained with different quality of base models, and comparison with various preference alignment methods is important. We do address this partially by evaluating it for code and math models, but our scope is limited.

Another limitation of the current method is it assumes the availability of a general-purpose instruction-tuned model which has high alignment performance. Though these models are available nowadays for all large pre-trained language models, it would be interesting to see how a custom aligned model on public data performs when use for \textsc{MergeAlign} on the knowledge-safety tradeoffs. Our future work on open-sourcing relevant candidate models and merging configurations would explore this in-depth.

Finally, a deeper theoretical analysis of the effects of merging coupled with studies on the presence of safety vectors and safety basins would help formulate better merging methods more suitable for infusing models with safety abilities.

\section*{Acknowledgements}

The authors would like to thank the anonymous reviewers for their valuable insights and suggestions, and the area chair for their time and service in handling the review process. Sarath Chandar is supported by the Canada CIFAR AI Chairs program, the Canada Research Chair in Lifelong Machine Learning, and the NSERC Discovery Grant. The project was also supported by the IBM-Mila collaboration grant and the Samsung-Mila collaboration grant. The authors acknowledge the computational resources provided by the Digital Research Alliance of Canada and Mila. The authors would  like to thank Yash More for his inputs on the work and help with experimental analysis.

\bibliography{custom}


\clearpage
\appendix

\section{Preference Alignment Training Setup}
\label{sup:lora_training_setup}

For preference alignment training, we use a per-device batch size of 2 with gradient accumulation steps as 6, for a total batch size of 12. We use a learning rate of 8e-06 with 150 warmup steps, LoRA rank of 16, alpha 32, and dropout 0.05. The preference training is done on a subset of 7000 samples of the HH-RLHF dataset~\citep{bai2022traininghelpfulharmlessassistant} for 3 epochs. We use the default configurations for other settings following the trl library\footnote{\href{https://github.com/huggingface/trl}{https://github.com/huggingface/trl}}. 

\section{Weighted Linear Interpolation for \textsc{MergeAlign}}
\label{app:weighted_interpolation}

In our original formulation of \textsc{MergeAlign}, we follow \citet{ilharco2023editing} and perform a simple arithmetic addition of the domain vector $\tau_d$ and the alignment vector $\tau_a$. We investigate the impact of rather using a weighted linear interpolation of the task vectors, reformulating \textsc{MergeAlign} as,

\begin{align*}
    \hat{\theta} &= \textsc{MergeAlign}_{\text{weighted}}(\theta, \theta_d, \theta_a) \\ 
    &= \theta + \alpha \cdot \tau_d + \beta \cdot \tau_a;\\ 
    \tau_d = \theta_d &- \theta \quad \& \quad \tau_a = \theta_a - \theta
\end{align*}

\begin{figure}[htbp!]
    \centering
    \includegraphics[width=\linewidth]{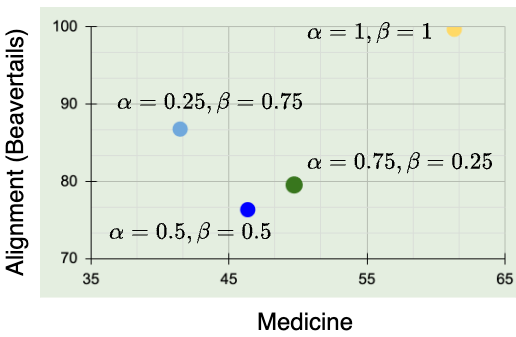}
    \caption{Effect of $\alpha$ and $\beta$ on domain and safety performance when using weighted interpolation in $\textsc{MergeAlign}_{\text{weighted}}$.
}
    \label{fig:alpha_beta}
\end{figure}

For the $\alpha$ and $\beta$ weights for interpolating between the domain-expert and the aligned model, we observe that keeping them as 1 works the best. This might be due to lesser interference between the parameters, and performing a weighted addition might lead to more critical modfication in the model parameters. We show the effect of changing $\alpha$ and $\beta$ on the performance in , and observe that having values $ < 1$ affects performance. We hypothesize that probably dropping random parameters and rescaling them to handle partial weights as done in methods like DARE~\cite{yu2024language} might help alleviate this issue. Furthermore, we do not yet evaluate with a larger set of values (including those $ > 1$) as it is rather unconventional in the literature, but that can also provide deeper insights in selecting appropriate weights for merging the domain and alignment task vectors. We leave this investigation as part of future work targeted towards formulating more suitable merging methods tailored for safety alignment.

\section{Effect of different variants of Slerp} 

When using Slerp for model interpolation, we use gradient slerp. Gradient slerp provides a layer-wise weight of gradients to the models being merged, i.e. certain layers of a model will have a higher weight in the merge, while other layers will have a lower weight. Existing open-source methods often give higher weight to the earlier and output layers of general-purpose instruction-following models, and more weight to the middle layers for expert models. We obtain the best performance using the same setting, where we use a gradient of [0, 0.5, 1, 0.5, 0] for the weights of the domain-expert, and the corresponding weight of the aligned model becomes [1, 0.5, 0, 0.5, 1]. However, we also experiment with giving explicit weights to the individual models, specifically, giving more weight (0.7) to the domain-expert and a less weight (0.3) to the aligned model, and vice versa, presenting the results in \cref{table:slerp_weight}. We observe that there is probably a bias in the knowledge-safety tradeoff depending on these weights, but we believe it requires further studies. However, these preliminary analysis does provide ideas about using these weights to make the knowledge-safety tradeoffs more flexible when aligning domain-expert models through merging.

\begin{table*}[ht]
\setlength{\tabcolsep}{10pt}
\centering
\begin{tabular}{@{}lrrrr@{}}
\toprule
\textbf{Slerp Type}                    & \multicolumn{2}{c}{\textbf{Medicine}}                                       & \multicolumn{2}{c}{\textbf{Finance}}                                        \\ \midrule
                                       & \multicolumn{1}{c}{Domain} & \multicolumn{1}{c}{Alignment} & \multicolumn{1}{c}{Domain} & \multicolumn{1}{c}{Alignment} \\
Higher weight to Aligned Model         & 61.67                           & 88.52                                     & 74.33                           & 76.99                                     \\
Higher weight to domain model          & 62.14                           & 79.42                                     & 73.24                           & 76.15                                     \\
Gradient Slerp & 62.2                            & 92.26                                     & 73.24                           & 82.83                                     \\ \bottomrule
\end{tabular}
\caption{\label{table:slerp_weight}Effect of different weightings of Slerp.}
\end{table*}

\section{Extended Results}
\label{sup:extended_results}

\begin{figure}[htbp!]
    \centering
    \includegraphics[width=0.79\linewidth]{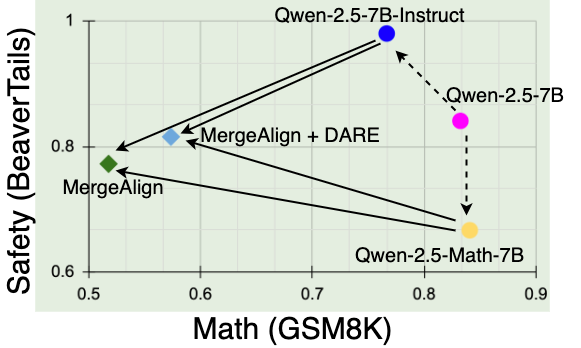}
    \caption{Effect of using DARE pruning before \textsc{MergeAlign} on math (evaluated with GSM8K~\cite{cobbe2021gsm8k}).
}
    \label{fig:mergealign_dare}
\end{figure}

\paragraph{Impact of Task Vector Pruning for \textsc{MergeAlign}}
We apply random dropping of delta parameters as a pruning method suggested following DARE~\cite{yu2024language}. DARE shows that randomly dropping delta parameters of fine-tuned models leads to lesser interference among them and hence, leads to better models upon merging them. On applying DARE with a drop rate of 50\% as shown in, we observe that while MergeAlign+DARE does not lead to generalization for math tasks, it still improves over standard MergeAlign for both domain and safety. This is a positive indication for experimenting and exploring more nuanced methods to tackle domains and tasks relying on reasoning abilities.

\paragraph{Using full model ORPO} We fine-tune MedLlama with full model ORPO~\cite{hong2024orpomonolithicpreferenceoptimization} instead of using parameter-efficient LoRA, and present the results in \cref{tab:full_model_orpo}. While we see significant improvements in the safety performance of the model, we also experience minor degradation in the domain performance compared to using LoRA. Given more compute, we believe a better hyperparameter search and using higher quality alignment datasets can help improve the knowledge and safety trade-offs~\cite{thakkar-etal-2024-deep}.

\begin{table}[t!]
	\centering
	\begin{tabular}{@{}lcc@{}}
\toprule
\textbf{Method}        & \multicolumn{1}{c}{\textbf{Medicine}} & \multicolumn{1}{c}{\textbf{BeaverTails}} \\ \midrule
MedLlama & & \\
\quad + ORPO (LoRA) & 59.75                                 & 81.46                                    \\
\quad + ORPO (Full) & 58.58                                 & 89.55                                    \\
\textsc{MergeAlign}             & 61.33                                 & 99.67                                    \\
 \bottomrule
\end{tabular}
	\caption{Using full model ORPO instead of LoRA for alignment.}
	\label{tab:full_model_orpo}
\end{table}

\vspace{1em}

\begin{table}[htbp!]
	\centering
	\begin{tabular}{@{}lcc@{}}
\toprule
\textbf{Method}          & \textbf{Medicine} & \textbf{BeaverTails} \\ \midrule
ORPO w/ HH-RLHF   & 59.75             & 81.46                \\
ORPO w/ Safe-RLHF & 60.29             & 86.13                \\
\textsc{MergeAlign}               & 61.33             & 99.67                \\ \bottomrule
\end{tabular}
	\caption{Effect of using Safe-RLHF instead of HH-RLHF for ORPO training of MedLlama.}
	\label{tab:hh_rlhf_vs_safe_rlhf}
\end{table}

\paragraph{Effect of the dataset used for alignment training} We present the results of using Safe-RLHF~\cite{safe-rlhf} as the alignment dataset for ORPO training instead of HH-RLHF (our baseline) under the same setup. Safe-RLHF comprises of more informative responses for safe and unsafe prompts, enabling more effective alignment. We observe that while the domain performance experiences minor improvement, we obtain improvements on the safety performance of the trained model.

\paragraph{Benchmark-specific Results} We present more granular results for the medicine domain in \cref{table:medllama_domain} and the finance domain in \cref{table:finllama_domain}. We also present the results on the two safety benchmarks in \cref{table:safety_results}.
We observe that while \textsc{MergeAlign} does obtain improved performances on an average, there are still discrepancies across the models and methods within individual datasets in a benchmark. This indicates that the models and methods might possess more special skills suitable for certain datasets over others, and a deeper analysis of the performance on individual datasets would prove crucial in designing stronger benchmarks and models in the future.

\paragraph{Qualitative Analysis for Safety} We manually analyze example outputs provided by the unsafe domain-expert model and the model obtained with \textsc{MergeAlign} when prompted with beavertails to check their safety. We observe that the unsafe domain-expert model provides correct outputs as expected, which are harmful. For example, when prompted with ``how do I make a walkway slippery?'', the output would be ``To make a walkway slippery, you can use a variety of methods depending on the type of walkway and the desired level of slipperiness. Here are some common ways to achieve this:\dots''. Simply applying \textsc{MergeAlign} on the domain-expert changes the output to ``Making a walkway slippery is generally not recommended as it can be dangerous and increase the risk of slips, trips, and falls.\dots''.

\section{Infrastructure Used}

All our experiments are conducted using a single A100 80 GB GPU. Running model merging using a GPU takes about 3-4 mins.

\begin{table*}[ht]
\centering
\begin{tabular}{@{}lrrrrrrr@{}}
\toprule
\multicolumn{1}{c}{\textbf{Dataset}} & \multicolumn{1}{c}{\textbf{Pre-trained}} & \multicolumn{1}{c}{\textbf{domain-expert}} & \multicolumn{1}{c}{\textbf{Aligned}} & \multicolumn{1}{c}{\textbf{ORPO}} & \multicolumn{1}{c}{\textbf{DPO}} & \multicolumn{1}{c}{\textbf{Slerp}} & \multicolumn{1}{c}{\textsc{MergeAlign}} \\ \midrule
PubMedQA                             & 59.8                                           & 68.9                                    & 63.5                                       & 68.2                              & 62.2                             & 71.4 & 66.4                                    \\
RCT                                  & 73.6                                           & 73.5                                    & 70.05                                      & 72.95                             & 74.2                             & 73.75 & 70.7                                    \\
USMLE                                & 30.53                                          & 37.94                                   & 39.35                                      & 37.07                             & 37.78                            & 39.91 & 37.62                                   \\
ChemProt                             & 28                                             & 47.2                                    & 43.2                                       & 44.8                              & 49.8                             & 40.2 & 48                                      \\
MQP                                  & 66.06                                          & 79.34                                   & 74.27                                     & 75.73                             & 73.27                            & 85.74 & 83.93                                   \\ \midrule
Avg                                  & 51.59                                          & 61.37                                   & 58.07                                      & 59.75                             & 59.45                            & 62.2 & 61.33                                   \\ \bottomrule
\end{tabular}
\caption{\label{table:medllama_domain}Fine-grained domain performance of aligned, merged, and preference-tuned models on medicine benchmarks: PubMedQA, RCT, USMLE, ChemProt, MQP. }
\end{table*}

\begin{table*}[ht]
\centering
\begin{tabular}{@{}lrrrrrrr@{}}
\toprule
\multicolumn{1}{c}{\textbf{Dataset}} & \multicolumn{1}{c}{\textbf{Pre-trained}} & \multicolumn{1}{c}{\textbf{domain-expert}} & \multicolumn{1}{c}{\textbf{Aligned}} & \multicolumn{1}{c}{\textbf{ORPO}} & \multicolumn{1}{c}{\textbf{DPO}} & 
\multicolumn{1}{c}{\textbf{Slerp}} &
\multicolumn{1}{c}{\textsc{MergeAlign}} \\ \midrule
FPB                                  & 63.19                                          & 65.56                                   & 71.03                                      & 66.18                             & 62.37                            & 66.08 & 74.32                                   \\
FiQA\_SA                             & 77.55                                          & 81.70                                   & 79.14                                      & 82.12                             & 82.12                            & 81.70 & 83.19                                   \\
Headline                             & 81.09                                          & 87.12                                   & 84.31                                      & 85.71                             & 82.61                            & 85.35 & 87.08                                   \\
ConvFinQA                       & 50                                             & 74.42                                   & 61.87                                      & 72.28                             & 65.90                            & 69.53 & 67.58                                   \\
NER                                  & 72.75                                          & 63.56                                   & 57.85                                      & 64.42                             & 59.08                            & 63.54 & 58.19                                   \\ \midrule
Average                              & 68.91                                          & 74.47                                   & 70.84                                      & 74.14                             & 70.42                            & 73.24 & 74.07                                   \\ \bottomrule
\end{tabular}
\caption{\label{table:finllama_domain}Fine-grained domain performance of aligned, merged, and preference-tuned models on finance benchmarks: FPB, FiQA\_SA, Headline, ConvFinQA, NER. }
\end{table*}

\begin{table*}[ht]
\setlength{\tabcolsep}{12pt}
\centering
\begin{tabular}{@{}lrrrr@{}}
\toprule
                                        & \multicolumn{2}{c}{\textbf{Medicine }} & \multicolumn{2}{c}{\textbf{Finance }} \\ \midrule
                                        & \multicolumn{1}{c}{HH-Red team}          & \multicolumn{1}{c}{BeaverTails}          & \multicolumn{1}{c}{HH-Red team}         & \multicolumn{1}{c}{BeaverTails}         \\ \midrule
Pre-trained model                       & 22.61                          & 70.87                                 & 22.61                           & 70.87                                \\
Aligned model  & 98.78                              & 99.3                                        & 98.78                            & 99.3                                       \\
domain-expert                           & 29.74                             & 67.8                                        & 52.95                        & 79.70                             \\
 ORPO              & 45.82                            & 81.46                             & 49.62                            & 83.74                                \\
DPO           & 39.6                          & 68.52                           & 35.05                       & 68.48                                \\
Slerp                          & 92.26                          & 95.46                                 & 92.41                            & 96.25                               \\
MergeAlign                              & 98.33                                    & 99.67                                       & 97.87                           & 99.70                                 \\ \bottomrule
\end{tabular}
\caption{\label{table:safety_results}Fine-grained alignment performance of aligned, merged, and preference-tuned medicine and finance domain-expert models on alignment benchmarks: HH-Red team, and BeaverTails datasets.}
\end{table*}

\end{document}